\documentclass{article}

\usepackage{amsmath,amsfonts,bm}

\def\eqref#1{equation~\ref{#1}}

\def\1{\bm{1}}

\def\rva{{\mathbf{a}}}

\def\rvs{{\mathbf{s}}}

\def\vq{{\bm{q}}}

\DeclareMathAlphabet{\mathsfit}{\encodingdefault}{\sfdefault}{m}{sl}
\SetMathAlphabet{\mathsfit}{bold}{\encodingdefault}{\sfdefault}{bx}{n}

\def\gA{{\mathcal{A}}}
\def\gB{{\mathcal{B}}}

\def\gD{{\mathcal{D}}}

\def\gH{{\mathcal{H}}}

\def\gL{{\mathcal{L}}}

\def\gR{{\mathcal{R}}}
\def\gS{{\mathcal{S}}}
\def\gT{{\mathcal{T}}}

\def\sH{{\mathbb{H}}}

\def\sR{{\mathbb{R}}}

\newcommand{\E}{\mathbb{E}}
\newcommand{\Ls}{\mathcal{L}}
\newcommand{\R}{\mathbb{R}}

\DeclareMathOperator*{\argmax}{arg\,max}
\DeclareMathOperator*{\argmin}{arg\,min}

\newcommand{\te}[1]{\texttt{#1}}

\newcommand\RotText[1]{\rotatebox{90}{\parbox{1.2cm}{\centering#1}}}

\usepackage{microtype}
\usepackage{graphicx}
\usepackage{subfigure}
\usepackage{booktabs} 

\usepackage{hyperref}

\usepackage[accepted]{icml2024}

\usepackage{amsmath}
\usepackage{amssymb}
\usepackage{mathtools}
\usepackage{amsthm}

\usepackage[capitalize,noabbrev]{cleveref}

\usepackage{tabularray} 
\usepackage{multirow}
\usepackage{multicol}
\usepackage{rotating} 
\usepackage{array}
\usepackage[hang]{footmisc}
\usepackage{pifont}
\usepackage{duckuments}
\usepackage{enumitem}

\usepackage[makeroom]{cancel}

\UseTblrLibrary{booktabs}

\theoremstyle{plain}
\newtheorem{theorem}{Theorem}[section]

\newtheorem{lemma}[theorem]{Lemma}

\theoremstyle{definition}

\theoremstyle{remark}

\newcommand{\PreserveBackslash}[1]{\let\temp=\\#1\let\\=\temp}
\newcolumntype{C}[1]{>{\PreserveBackslash\centering}p{#1}}
\newcolumntype{R}[1]{>{\PreserveBackslash\raggedleft}p{#1}}
\newcolumntype{L}[1]{>{\PreserveBackslash\raggedright}p{#1}}

\usepackage[textsize=tiny]{todonotes}

\icmltitlerunning{Q-value Regularized Transformer for Offline Reinforcement Learning}

\begin{document}

\twocolumn[
\icmltitle{Q-value Regularized Transformer for Offline Reinforcement Learning}




\begin{icmlauthorlist}
\icmlauthor{Shengchao Hu}{sjtu,pjlab}
\icmlauthor{Ziqing Fan}{sjtu,pjlab}
\icmlauthor{Chaoqin Huang}{sjtu,pjlab}
\icmlauthor{Li Shen}{zs,jd}
\icmlauthor{Ya Zhang}{sjtu,pjlab}
\icmlauthor{Yanfeng Wang}{sjtu,pjlab}
\icmlauthor{Dacheng Tao}{ntu}
\end{icmlauthorlist}

\icmlaffiliation{sjtu}{Shanghai Jiao Tong University, China}
\icmlaffiliation{pjlab}{Shanghai AI Laboratory, China}
\icmlaffiliation{zs}{Sun Yat-sen University, China}
\icmlaffiliation{ntu}{Nanyang Technological University, Singapore}
\icmlaffiliation{jd}{JD Explore Academy, China}

\icmlcorrespondingauthor{Li Shen}{mathshenli@gmail.com}

\icmlkeywords{Machine Learning, ICML}

\vskip 0.3in
]



\printAffiliationsAndNotice{}  

\begin{abstract}
    Recent advancements in offline reinforcement learning (RL) have underscored the capabilities of Conditional Sequence Modeling (CSM), a paradigm that learns the action distribution based on history trajectory and target returns for each state.
    However, these methods often struggle with stitching together optimal trajectories from sub-optimal ones due to the inconsistency between the sampled returns within individual trajectories and the optimal returns across multiple trajectories.
    Fortunately, Dynamic Programming (DP) methods offer a solution by leveraging a value function to approximate optimal future returns for each state, while these techniques are prone to unstable learning behaviors, particularly in long-horizon and sparse-reward scenarios.
    Building upon these insights, we propose the Q-value regularized Transformer (QT), which combines the trajectory modeling ability of the Transformer with the predictability of optimal future returns from DP methods.
    QT learns an action-value function and integrates a term maximizing action-values into the training loss of CSM, which aims to seek optimal actions that align closely with the behavior policy. 
    Empirical evaluations on D4RL benchmark datasets demonstrate the superiority of QT over traditional DP and CSM methods, highlighting the potential of QT to enhance the state-of-the-art in offline RL.
\end{abstract}
\section{Introduction}

Offline reinforcement learning (RL) aims at learning effective policies entirely from previously collected data without interacting with the environment \citep{fujimoto2019off}.
Recent advancements in offline RL have taken a new perspective on the problem, departing from conventional methods for offline RL that concentrate on policy regularization \citep{BEAR, fujimoto2019off} or conservatism for value function approximation \citep{kostrikov2021offline, CQL}. 
Instead, the problem is viewed as a generic Conditional Sequence Modeling (CSM) task \citep{DT, TT}, where past experiences consisting of state-action-reward triplets are input to Transformer \citep{transformer}. 
The model generates a sequence of action predictions using a goal-conditioned policy, effectively converting offline RL to a supervised learning problem.
This approach relaxes the MDP assumption by considering multiple historical steps to predict an action, allowing the model to be capable of handling long sequences and avoid stability issues associated with bootstrapping \citep{srivastava2019training, kumar2019reward}. 

However, the CSM approach fails to achieve the stitching property desired in offline RL, which involves synthesizing optimal trajectories from sub-optimal ones \citep{fu2020d4rl}.
The primary challenge lies in the inconsistency between sampled target returns and the optimal returns from actions, as high-return trajectories might not reflect superior actions but rather fortunate circumstances \citep{wang2023critic}. 
CSM associates the return-to-go (RTG) token value with individual trajectories, overlooking the stochastic nature of state transitions and optimal future returns that span across different trajectories \citep{paster2022you}.
Additionally, the intrinsic uncertainty and approximation errors in behavior policies further exacerbate the inconsistency, leading to inferior performance in stitching tasks, particularly when dealing with sub-optimal data \citep{wang2023critic}.

Fortunately, conventional Dynamic Programming methods (Q-learning\footnote{In this paper, the terms \textit{Q-learning} and \textit{Dynamic Programming (DP)} will be used interchangeably to refer to any RL algorithm that relies on the Bellman-backup operation.}) provide a robust solution to handle this inconsistency.
By treating each timestep individually and backpropagating optimal future returns for each state, these methods enable agents to select actions that maximize long-term returns. 
However, these techniques are prone to unstable learning behaviors, particularly in long-horizon and sparse-reward scenarios \citep{QDT}.
While the conceptual integration of Q-learning with CSM is straightforward, developing a framework that effectively unites their strengths and overcomes their limitations poses a significant challenge.
QDT \citep{QDT} takes the first attempt to combine these two methods by learning a conservative value function to relabel the RTG values while remaining other components the same as DT \citep{DT}.
This approach seeks to enhance stitching capability by incorporating augmented trajectories into the training dataset. 
However, empirical evaluations suggest that while it may alleviate some issues, it still struggles with unmatched RTG values during inference arising from trajectory-level modeling \citep{wang2023critic}, often achieving results comparable to but not exceeding existing methods (Figure \ref{fig:toy}).

Building upon these insights, we propose the Q-value regularized Transformer (QT), which combines the trajectory modeling ability with the predictability of optimal future returns from DP methods.
Our policy is based on a Transformer structure, with an objective loss comprising two components: 1) a conditional behavior cloning term that aligns the Transformer's action sampling with the training set's distribution, and 2) a policy improvement term for selecting high-reward actions according to the learned Q-value.
This hybrid structure offers multiple advantages. 
First, the trajectory prediction loss serves as an effective distribution-matching technique, functioning as a robust, sample-based policy regularization method, thus eliminating the need for additional behavior cloning. 
Second, the integration of policy improvement facilitates the identification and prioritization of higher-reward actions as per Q-values, ensuring that the expected returns of sampled actions align with the optimal returns. 
Third, the amalgamation of these two losses achieves a balance between selecting optimal actions and maintaining fidelity to the behavior policy, which mitigates the risk of preferring out-of-distribution actions with overestimated values, leading to enhanced performance.

In summary, our contributions are three-fold\footnote{Our code is available at: \url{https://github.com/charleshsc/QT}}: 
\begin{itemize}[leftmargin=*]
    \item QT, a new offline RL algorithm that leverages Transformer to do precise policy regularization and Q-value regularization to align the expected returns of sampled actions with the optimal returns.
    \item QT aims to seek optimal actions that align closely with the behavior policy, ensuring robust stitching capability and effective trajectory modeling in scenarios characterized by long horizons and sparse rewards.
    \item We test QT on the D4RL benchmark tasks and demonstrate the superiority of QT over traditional DP and CSM methods, highlighting the potential of QT to enhance the state-of-the-art in offline RL.
\end{itemize}
\section{Preliminary}
\label{sec:preliminary}

\subsection{Offline Reinforcement Learning}

The goal of RL is to learn a policy $\pi_{\theta}(\rva | \rvs)$ maximizing the expected cumulative discounted rewards $\E[\sum_{t=0}^{\infty} \gamma^t \gR(\rvs_t, \rva_t)]$ in a Markov decision process (MDP), which is a six-tuple $(\gS, \gA, \gT, \gR, \gamma, d_0)$, with state space $\gS$, action space $\gA$, environment dynamics $\gT(\rvs' | \rvs, \rva) : \gS \times \gS \times \gA \rightarrow [0,1]$, reward function $\gR: \gS \times \gA \rightarrow \R$, discount factor $\gamma \in [0, 1)$, and initial state distribution $d_0$ \citep{sutton2018reinforcement}.
The action-value or Q-value of a policy $\pi$ is defined as $Q^{\pi}(\rvs_t, \rva_t) = \E_{\rva_{t+1}, \rva_{t+2} \dots \sim \pi} [\sum_{i=0}^{\infty} \gamma^i \gR(s_{t+i}, a_{t+i})]$.
In the offline setting \citep{levine2020offline}, instead of the online environment, a static dataset $\gD = \{(\rvs, \rva, \rvs', r)\}$, collected by a behavior policy $\pi_{\beta}$, is provided.
Offline RL algorithms learn a policy entirely from this static offline dataset $\gD$, without online interactions with the environment.

\subsection{Rethinking Stitching in CSM}
\label{sec:rethink}

To address the stitching ability of CSM, alternative approaches have been proposed.
For example, EDT \citep{wu2023elastic} and CGDT \citep{wang2023critic} optimize the trajectory by dynamically filtering the optimal trajectory according to the learned value estimator;
ESPER \citep{paster2022you} clusters trajectories and utilizes the average cluster returns as conditions for the policy; 
DoC \citep{yang2022dichotomy} conditions the policy on a latent representation of future trajectories, achieved by minimizing mutual information.
Incorporating probabilistic statistics from multiple trajectories offers a promising solution for sub-optimal data, which guides policy behaviors with learned estimated returns from the entire distribution of future trajectories.
Although these methods exhibit effectiveness in stitching ability, they often necessitate complex objectives for representation learning and additional steps such as statistics, thereby complicating and burdening the training process.

\begin{figure}[!t]
    \centering
    \subfigure[]{
        \includegraphics[width=0.45\linewidth]{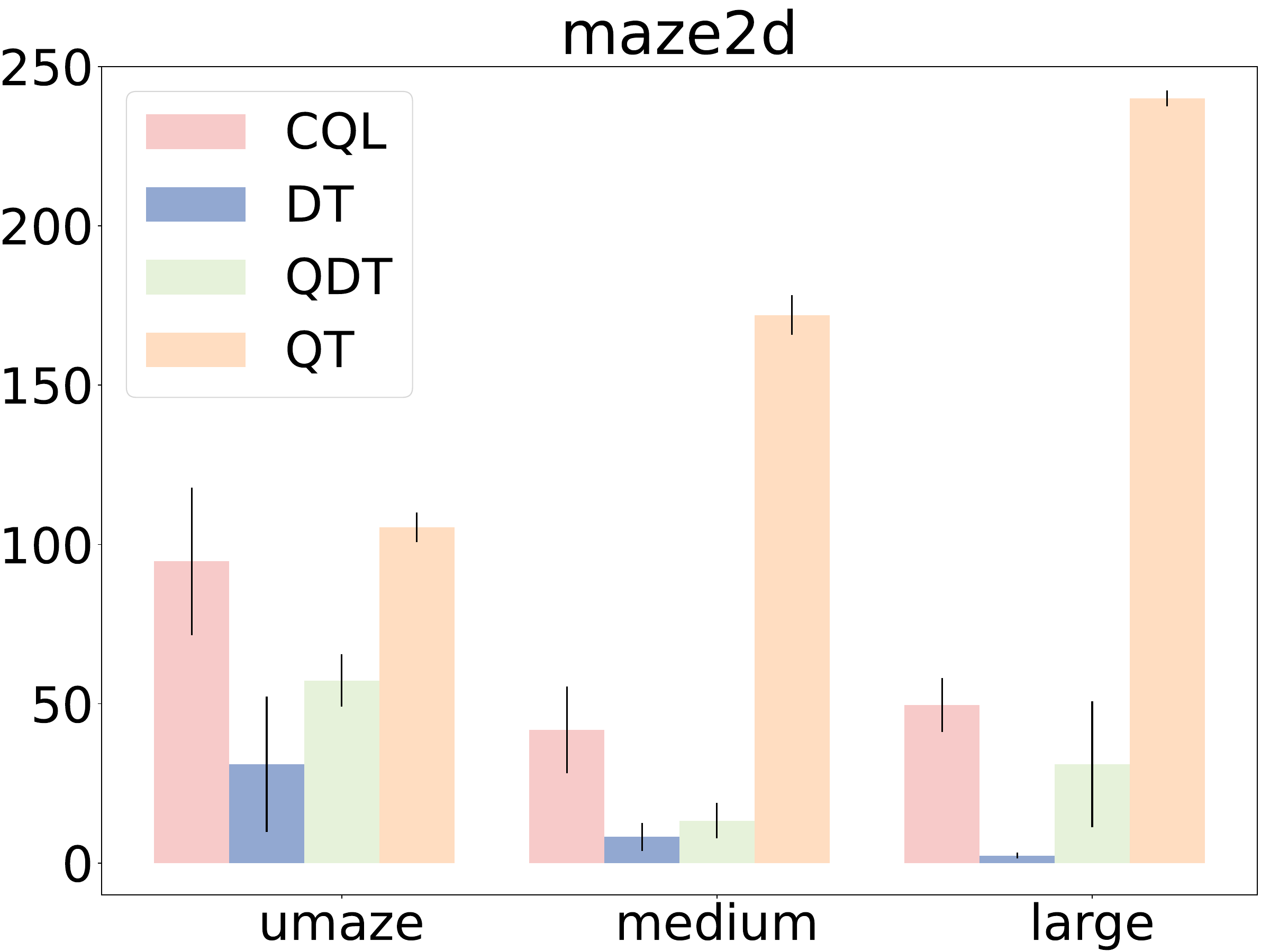}
        \label{fig:simple_exp}
    }
    \subfigure[]{
	\includegraphics[width=0.45\linewidth]{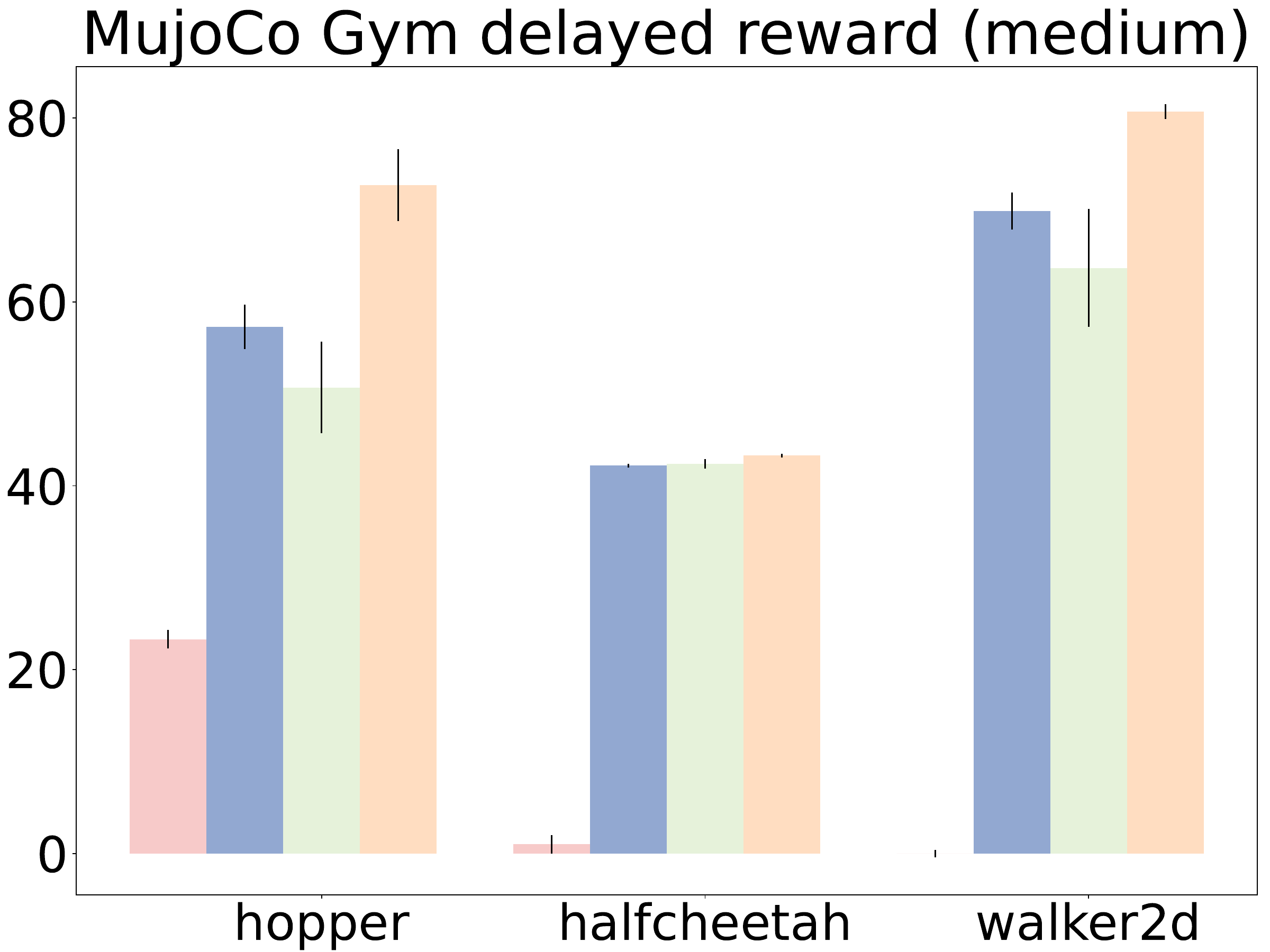}
        \label{fig:initia}
    }
    \vspace{-0.4cm}
    \caption{Evaluation results for CQL, DT, QDT, and QT in the Maze2D tasks (a) and MuJoCo Gym delayed reward (medium) tasks (b). 
    The results show that DT fails to effectively stitch trajectories and CQL under-performs in sparse reward scenarios (delayed reward). 
    QDT yields consistent yet intermediate results across all environments, while QT consistently secures the top performance across all tested environments, showcasing its superiority.
    }
    \label{fig:toy}
\vspace{-0.2cm}
\end{figure}

Other approaches exploit the capabilities of Q-learning, which propagates optimal future returns backward for each state, considering each time step individually, thereby effectively stitching the optimal trajectory from sub-optimal data.
QDT \citep{QDT} takes the first attempt to combine these two methods by learning a conservative value function to relabel the RTG tokens in the dataset, keeping other components aligned with DT \citep{DT}. 
However, such adaptations essentially constitute simple data augmentation, incorporating "stitched" trajectories into the training set but continuing to encounter unmatched RTG values during inference due to trajectory-level modeling \citep{wang2023critic}, thereby failing to consistently exceed existing benchmarks.
In contrast, QT employs the n-step Bellman equation to approximate the Q-value function based on sequence history. This Q-value function is then integrated into policy improvement to select high-reward actions while retaining the original DT loss for policy regularization.
Such an approach not only empowers the CSM with the stitch ability but also keeps its original trajectory modeling ability important for the sparse-reward scenario.
To substantiate this, we compared QT and QDT across various scenarios, including stitching ability scenarios like Maze2D, and sparse reward scenarios like MujoCo Gym with delayed rewards.
The results, as illustrated in Figure \ref{fig:toy}, show that QT consistently achieves superior performance, while QDT's results are intermediate, failing to exceed existing methodologies (more details are presented in Section \ref{sec:ab}).

\section{Methodology}
\label{sec:method}
We present a method that combines the trajectory modeling ability of Transformer with the predictability of optimal future returns from DP methods, thereby constructing a robust algorithm suitable for offline RL problems. 
Initially, we detail the application of the Conditional Transformer Policy as an expressive policy framework for behavior cloning. 
Subsequently, we describe the incorporation of a Q-value module into the training phase of our transformer policy, with the behavior cloning term serving as a policy regularization mechanism.
Finally, we illustrate how to do the inference with the learned Q-value functions.

\begin{algorithm*}[tb]
   \caption{QT: Q-value regularized Transformer}
   \label{alg:QT}
\begin{algorithmic}
   \STATE {\bfseries Input:} Sequence horizon $K$, offline datasets $\gD$, coefficient $\rho$, a set of candidate return-to-go $\{\hat{r}_0^0, \hat{r}_0^1, \dots, \hat{r}_0^m \}$.
   \STATE Initialize policy network $\pi_{\theta}$, critic networks $Q_{\phi_1}$ and $Q_{\phi_2}$, and target networks $\pi_{\theta'}, Q_{\phi_1'}$ and $Q_{\phi_2'}$.
   \STATE {\color{gray}\te{// Train the QT}}
   \FOR{$t=1$ {\bfseries to} $T$}
   \STATE Sample sequence transition mini-batch $\gB = \{ (\hat{r}_j, \rvs_j, \rva_j, r_j)_{j=t}^{t+K} \} \sim \gD$.
   \STATE {\color{gray}\te{// Q-value function learning}}
   \STATE Sample $\hat{\rva}_{t+K} \sim \pi_{\theta'} (\hat{\rva}_{t+K} | \hat{r}_{t : t+K}, \rvs_{t : t+K}, \rva_{t : t+K-1})$.
   \STATE Update $Q_{\phi_1}$ and $Q_{\phi_2}$ by Equation \ref{eq:Q_update}. 
   \STATE {\color{gray}\te{// Policy learning}}
   \FOR{$i=1$ {\bfseries to} $K$}
   \STATE Sample $\hat{\rva}_{t+i} \sim \pi_{\theta} (\hat{\rva}_{t+i} | \hat{r}_{t:t+i}, \rvs_{t:t+i}, \rva_{t:t+i-1})$ in an auto-regressive way.
   \ENDFOR.
   \STATE Update policy by minimizing Equation 
   \ref{eq:final_update}.
   \STATE {\color{gray}\te{// Update target networks}}
   \STATE $\theta' = \rho \theta' + (1-\rho)\theta, \phi'_i = \rho \phi_i' + (1-\rho) \phi_i$ for $i = \{1 ,2\}$.
   \ENDFOR.
   \STATE {\color{gray}\te{// Inference with QT}}
   \STATE Given multiple target return-to-go choice $\hat{r}^{1:m}_0$ and initial state $s_0$.
   \REPEAT
   \STATE Sample multiple actions with different return-to-go $\hat{\rva}^i_t = \pi_{\theta} (\hat{\rva}^i_t | \hat{r}^i_{t-K+1:t}, \rvs_{t-K+1:t}, \rva_{t-K+1:t-1})$ for $i={1, \dots, m}$.
   \STATE Compute Q value with candidate state-action pair $(\rvs_t, \hat{\rva}_t^i)$ for $i={1, \dots, m}$.
   \STATE Sample the action $\rva_t$ from action set $\{ \hat{\rva}_t^i \}_{i=1}^m$ with the max Q value by Equation \ref{eq:inf}.
   \STATE Execute the action $\rva_t$ and collect the reward $r_t$ and next state $\rvs_{t+1}$.
   \STATE Update current return-to-go $\hat{r}^i_{t+1} = \hat{r}^i_{t} - r_t$ for $i={1, \dots, m}$.
   \UNTIL{$Done$ is $true$.}
\end{algorithmic}
\end{algorithm*}

\subsection{Conditional Transformer Policy}
\label{sec:CTP}

Transformer \citep{vaswani2017attention}, extensively studied in NLP \citep{bert} and CV \citep{ViT}, has also been explored in RL using the CSM pattern \citep{hu2022transforming}. 
Unlike the majority of prior RL approaches that estimate value functions or compute policy gradients, DT \citep{DT} outputs desired future actions from the history sequence, encompassing multiple state $\rvs_t$, action $\rva_t$, and return-to-go $\hat{r}_t$ tuples.
The return-to-go token quantifies the cumulative reward from the current time step to the end of the episode.
During training with offline collected data, DT processes a trajectory sequence $\tau_t$ in an auto-regressive manner which encompasses the most recent K-step historical context:
\begin{equation}
\label{eq:input}
    \tau_t = (\hat{r}_{t-K+1}, \rvs_{t-K+1}, \rva_{t-K+1}, \dots,  \hat{r}_{t}, \rvs_{t}, \rva_{t}).
\end{equation}

The prediction head associated with a state token $\rvs_t$ is trained to predict the corresponding action $\rva_t$.
Regarding continuous action spaces, the training objective is to minimize the mean-squared loss:
\begin{equation}
\label{eq:DTLoss}
\small
   \Ls_{DT} = \mathbb{E}_{\tau_t \sim \gD} \left[ \frac{1}{K} \sum_{i=t-K+1}^t (\rva_i - \pi(\tau_t)_i )^2 \right],
\end{equation}
where $\pi(\tau_t)_i$ denotes the $i$-th action output of the Transformer policy in an auto-regressive manner.

\begin{theorem}
\label{thm:onlydt}
Consider an MDP, behavior policy $\beta$, and decision transformer $\pi$ with condition function $f$. 
Assume the $\epsilon$-near determinism of the MDP, where $P( r \neq \gR(\rvs,\rva) ~or~ \rvs' \neq \gT(\rvs,\rva) | \rvs,\rva) \leq \epsilon$ at all $\rvs, \rva$ for some functions $\gT$ and $\gR$.
Let $g(\tau) = \sum_{t=1}^\gH r_t$, when $P_{\beta} (g(\tau) = f(\rvs_1) |\rvs_1 ) \geq \alpha_{f}$ for all initial states $\rvs_1$, we have:
\begin{equation}
\small
    \E_{\tau \sim \beta}[g(\tau)] - \E_{\tau \sim \pi_f} [g(\tau)] \leq \epsilon (\frac{1}{\alpha_f} + 2) \gH^2,
\end{equation}
where $\gH$ is the horizon of the MDP.
\end{theorem}

Theorem \ref{thm:onlydt} demonstrates that training with the DT loss $\mathcal{L}_{DT}$ leads to the gradual convergence of the generated policy towards the behavior policy $\beta$. 
This convergence, however, imposes a constraint that restricts the generated policy from exceeding the performance of the behavior trajectories present in the offline dataset $\gD$.
Moreover, training exclusively with the DT loss $\Ls_{DT}$ restricts the stitching ability, resulting in a policy predominantly biased towards actions observed in the training trajectories \citep{paster2022you}. 
Due to limited space, the proof of this theorem, as well as other results, are provided in the Appendix \ref{sec:proof}.

\subsection{Training with Q-value Regularization}
To address the stitching challenge and develop a policy capable of aligning the expected returns of sampled actions with the optimal returns, we employ the Q-value module.

The Q-value function is learned conventionally, minimizing the Bellman operator \citep{fujimoto2019off} and employing the double Q-learning technique \citep{hasselt2010double}.
We construct two Q-networks, $Q_{\phi_1}, Q_{\phi_2}$, along with their respective target networks, $Q_{\phi_1'}, Q_{\phi_2'}$ and target policy $\pi_{\theta'}$.
Given that the input to the transformer policy includes trajectory history, we opt for the n-step Bellman equation to estimate the Q-value function.
This choice is premised on its demonstrated improvement over the 1-step approximation \citep{sutton2018reinforcement}.
The optimization of $\phi_i$ for $i=\{1,2\}$ is carried out by minimizing following equation:
\begin{align}
\small
\label{eq:Q_update}
    &\E_{\tau_t \sim \gD, \hat{\rva}_t \sim \pi_{\theta'}}  
      \sum_{m=t-K+1}^{t-1} 
    \Big|\Big| \hat{Q}_m - Q_{\phi_i}(\rvs_m, \rva_m) \Big|\Big|^2, \\ \nonumber
     &\text{where}~~ \hat{Q}_m = \sum_{j=m}^{t-1} \gamma^{j-m} r_j
     + \gamma^{t-m} \min_{i=1,2} Q_{\phi_i'} (\rvs_t, \hat{\rva}_t),
\end{align}
where $\gamma$ is the discount factor and $\hat{\rva}_t$ denotes the predicted action output by the target model $\pi_{\theta'}$.

To enhance the policy, we integrate a Q-value module during the training phase, enabling the preferential sampling of high-value actions. 
The final policy learning objective emerges as a linear combination of policy regularization and policy improvement elements:
\begin{align}
\label{eq:final_update}
    \pi &= \argmin_{\pi_{\theta}} \left\{\gL(\theta)
    := \gL_{DT} (\theta) + \gL_{Q} (\theta)\right\} \\
    &= \! \argmin_{\pi_{\theta}}  \gL_{DT}(\theta) \!-\! \alpha \cdot \E_{\tau_t \sim \gD} \E_{(\rvs_i, \rva_i) \sim \tau_t} Q_{\phi}(\rvs_i, \pi(\tau_t)_i). \!\nonumber
\end{align}
Considering the variation in the scale of the Q-value function across different offline datasets, we adopt a normalization technique from \citet{TD3BC}.
We define $\alpha$ as $\alpha = \frac{\eta}{\E_{\tau_t \sim \gD} \E_{(\rvs, \rva) \sim \tau_t} \left[ |Q_{\phi} (\rvs, \rva)| \right]}$, where $\eta$ is a hyper-parameter that mediates the balance between the two loss terms. 
Notably, the Q-value in the denominator serves exclusively for normalization and is not subject to differentiation.

Furthermore, we affirm the efficacy of Equation \ref{eq:final_update} from a theoretical standpoint as delineated in Theorem \ref{thm:qdt}, suggesting that the learned final policy is anticipated to consistently outperform the behavior policy in terms of the value function. 
Specifically, it highlights how the Q-value regularization enhances the policy by enabling preferential sampling of high-value actions, aligning the learning process more closely with optimal returns.
This implicitly ensures an improvement over the baseline behavior policy $\beta$.

\begin{theorem}
\label{thm:qdt}
Let $\pi^*$ be the optimal policy of Equation \ref{eq:final_update}. For any $\rvs \in \gS$, we have that $V^{\pi^*}(\rvs) \geq V^{\beta}(\rvs)$ and $\pi^*(\rva|\rvs) = 0$ given $\beta(\rva|\rvs) = 0$.
\end{theorem}

\begin{table*}[!ht]
 \caption{The performance of QT and SOTA baselines on D4RL Gym, Adroit, Kitchen, Maze2D, and AntMaze tasks. Results for QT correspond to the mean and standard errors of normalized scores over 30 random rollouts (3 independently trained models and 10 trajectories per model) for all tasks, which generally exhibit low variance in performance. Our method outperforms all prior methods by a clear margin in almost all domains, including the conventional Q-learning algorithms and CSM methods. }
 \vspace{0.1cm}
    \label{tab:main_result}
    \centering
    \resizebox{\textwidth}{!}{
    \begin{tblr}{
    colspec = {l||*{6}{c}|*{6}{c}|c},
    row{1, 12, 20, 24, 29} = {font=\bfseries}
    }
    \toprule
        Gym Tasks & CQL & IQL & BCQ & BEAR & TD3+BC & MoRel & BC & DD &  DT & StAR & GDT & CGDT & QT \\
        \hline
        halfcheetah-medium-expert-v2 & 91.6 & 86.7 & 69.6 & 53.4 & 90.7 & 53.3 & 55.2 & 90.6 &  86.8 & 93.7 & 93.2 & 93.6 & \textbf{96.1} {\small $\pm$ 0.2}\\ 
        hopper-medium-expert-v2 & 105.4 & 91.5 & 109.1 & 96.3 & 98.0 & 108.7 & 52.5 & 111.8 & 107.6 & 111.1 & 111.1 & 107.6 & \textbf{113.4} {\small $\pm$ 0.4} \\ 
        walker2d-medium-expert-v2 & 108.8 & 109.6 & 67.3 & 40.1 & 110.1 & 95.6 & 107.5 & 108.8 & 108.1 & 109.0 & 107.7 & 109.3 & \textbf{112.6} {\small $\pm$ 0.6} \\ 
        halfcheetah-medium-v2 & 49.2 & 47.4 & 41.5 & 41.7 & 48.4 & 42.1 & 42.6 & 49.1 & 42.6 & 42.9 & 42.9 & 43.0 & \textbf{51.4} {\small $\pm$ 0.4}\\
        hopper-medium-v2 & 69.4 & 66.3 & 65.1 & 52.1 & 59.3 & 95.4 & 52.9 & 79.3 & 67.6 & 59.5 & 77.1 & \textbf{96.9} & \textbf{96.9} {\small $\pm$ 3.1} \\ 
        walker2d-medium-v2 & 83.0 & 78.3 & 52.0 & 59.1 & 83.7 & 77.8 & 75.3 & 82.5 & 74.0 & 73.8 & 76.5 & 79.1 & \textbf{88.8} {\small $\pm$ 0.5}\\
        halfcheetah-medium-replay-v2 & 45.5 & 44.2 & 34.8 & 38.6 & 44.6 & 40.2 & 36.6 & 39.3 & 36.6 & 36.8 & 40.5 & 40.4 & \textbf{48.9} {\small $\pm$ 0.3}\\
        hopper-medium-replay-v2 & 95.0 & 94.7 & 31.1 & 33.7 & 60.9 & 93.6 & 18.1 & 100.0 & 82.7 & 29.2 & 85.3 & 93.4 & \textbf{102.0} {\small $\pm$ 0.2} \\
        walker2d-medium-replay-v2 & 77.2 & 73.9 & 13.7 & 19.2 & 81.8 & 49.8 & 32.3 & 75.0 & 79.4 & 39.8 & 77.5 & 78.1 & \textbf{98.5} {\small $\pm$ 1.1} \\
        \hline 
        \textbf{Average} & 80.6 & 77.0 & 53.8 & 48.2 & 75.3 & 72.9 & 52.6 & 81.8 & 76.2 & 66.2 & 79.1 &  82.4 &  \textbf{89.8} \\
    \bottomrule
    \toprule
        Adroit Tasks & CQL & IQL & BCQ & BEAR & O-RL & MoRel & BC & DD & D-QL & DT & StAR & GDT & QT \\ \hline
        pen-human-v1 & 37.5 & 71.5 & 66.9 & -1.0 & 90.7 & -3.2 & 63.9 & 66.7 & 72.8 & 79.5 & 77.9 & 92.5 & \textbf{129.6} {\small $\pm$ 4.6} \\
        hammer-human-v1 & 4.4 & 1.4 & 0.9 & 0.3 & 0.2 & 2.7 & 1.2 & 1.9  & 0.2 & 3.7 & 3.7 & 5.5 & \textbf{35.6} {\small $\pm$ 7.0}\\
        door-human-v1 & 9.9 & 4.3 & -0.05 & -0.3 & -0.1 & 2.2 & 2.0 & 2.8 & 0.0 & 14.8 & 1.5 & 20.6 & \textbf{28.7} {\small $\pm$ 2.4} \\
        pen-cloned-v1 & 39.2 & 37.3 & 50.9 & 26.5 & 60 & -0.2 & 37.0 & 42.8  & 57.3 & 75.8 & 33.1 & 86.2 & \textbf{125.0} {\small $\pm$ 2.8}\\ 
        hammer-cloned-v1 & 2.1 & 2.1 & 0.4 & 0.3 & 2.0 & 2.3 & 0.6 & 1.7  & 3.1 & 3.0 & 0.3 & 8.9 & \textbf{23.0} {\small $\pm$ 2.3}\\
        door-cloned-v1 & 0.4 & 1.6 & 0.01 & -0.1 & 0.4 & 2.3 & 0.0 & 1.3 & 0.0 & 16.3 & 0.0 & 19.8 & \textbf{20.6} {\small $\pm$ 1.7} \\
        \hline
        \textbf{Average} & 15.6 & 19.7 & 19.8 & 4.3 & 25.5 & 1.0 & 17.5 & 19.5 & 22.2 & 32.2 & 19.4 & 38.9 & \textbf{60.4}  \\ 
    \bottomrule
    \toprule
        Kitchen Tasks & CQL & IQL & BCQ & BEAR & TD3+BC & O-RL & BC & DD & D-QL & DT & StAR & GDT & QT \\ \hline
        kitchen-complete-v0 & 43.8 & 62.5 & 8.1 & 0.0 & 0.0 & 2.0  & 65.0 & 65.0  & \textbf{84.0} & 50.8 & 40.8 & 43.8 & 81.7 {\small $\pm$ 1.2} \\
        kitchen-partial-v0 & 49.8 & 46.3 & 18.9 & 13.1 & 0.0 & 35.5 & 33.8 & 57.0 & 60.5 & 57.9 & 12.3 & 73.3 & \textbf{75.0} {\small $\pm$ 0.1} \\
        \hline
        \textbf{Average} & 46.8 & 54.4 & 13.5 & 6.6 & 0.0 & 18.8 & 51.5 & 61 & 72.3 & 54.4 & 26.6 & 58.6 & \textbf{78.4}   \\ 
    \bottomrule
    \toprule
        Maze2D Tasks & CQL & IQL & BCQ & BEAR & TD3+BC & COMBO & BC & Diffuser & DD & DT & GDT & QDT & QT \\ \hline
        maze2d-umaze-v1 & 94.7 & 42.1 & 49.1 & 65.7 & 14.8 & 76.4 & 88.9 & 113.9 & \textbf{116.2} & 31.0 & 50.4  & 57.3 & 105.4 {\small $\pm$ 4.7} \\
        maze2d-medium-v1 & 41.8 & 34.9 & 17.1 & 25.0 & 62.1 & 68.5 & 38.3 & 121.5 & 122.3 & 8.2 & 7.8  & 13.3 & \textbf{172.0} {\small $\pm$ 6.2}\\
        maze2d-large-v1 & 49.6 & 61.7 & 30.8 & 81.0 & 88.6 & 14.1 & 1.5 & 123.0 & 125.9 & 2.3 & 0.7  & 31.0 & \textbf{240.1} {\small $\pm$ 2.5}\\
        \hline
        \textbf{Average} & 62.0 & 46.2 & 32.3 & 57.2 & 55.2 & 53.0 & 42.9 & 119.5 & 121.5 & 13.8 & 19.6 & 33.9 & \textbf{172.5} \\
    \bottomrule
    \toprule
        AntMaze Tasks & CQL & IQL & BCQ & BEAR & TD3+BC & O-RL & BC & DD & D-QL & DT & StAR & GDT & QT \\ \hline
        antmaze-umaze-v0 & 74.0 & 87.5 & 78.9 & 73.0 & 78.6 & 64.3 & 54.6 & 73.1 & 93.4 & 59.2 & 51.3  & 76.0 & \textbf{96.7} {\small $\pm$ 4.7} \\
        antmaze-umaze-diverse-v0 & 84.0 & 62.2 & 55.0 & 61.0 & 71.4 & 60.7 & 45.6 & 49.2 & 66.2 & 53.0 & 45.6 & 69.0 & \textbf{96.7} {\small $\pm$ 4.7}\\
        antmaze-medium-diverse-v0 & 53.7 & 70.0 & 0.0 & 8.0 & 3.0 & 0.0 & 0.0 & 24.6 & \textbf{78.6} & 0.0 & 0.0 & 6.0 & 59.3 $\pm$ 0.9 \\
        antmaze-large-diverse-v0 & 14.9 & 47.5 & 2.2 & 0.0 & 0.0 & 0.0 & 0.0 & 7.5 & \textbf{56.6} & 0.0 & 0.0 & 0.0 & 53.3 $\pm$ 4.7\\
        \hline
        \textbf{Average} & 56.7 & 66.8 & 34.0 & 57.2 & 38.3 & 31.3 & 25.1 & 61.2 & 73.7 & 28.1 & 24.2 & 37.8 & \textbf{76.5} \\
    \bottomrule
    \end{tblr}}
    \vspace{-2mm}
\end{table*}

\subsection{Inference with Q-value Module}

Instead of carefully designing the return-to-go token value in the previous conditional transformer policy, which needs more trials and tuning to find the best value, we sample multiple candidate return-to-go tokens $\{\hat{r}_0^0, \hat{r}_0^1, \dots, \hat{r}_0^m \}$ and simultaneously output actions in accordance with different return-to-go values.
Then we resort to the learned Q-value function to preferentially sample actions with high returns, which could be formulated as:
\begin{align}
\label{eq:inf}
    &\argmax_{\hat{\rva}_t^i} ~~ Q_{\phi'}( \rvs_t, \hat{\rva}_t^i), \\
    \text{where} \quad \hat{\rva}_t^i = &\pi(\hat{r}^i_{t-K+1:t}, \rvs_{t-K+1:t},\rva_{t-K+1:t-1} ))\nonumber.
\end{align}
This process is highly parallelizable. By assigning different RTG values to each batch, we can leverage GPU capabilities to concurrently generate multiple action sequences, thereby minimizing additional computational overhead.
Corresponding ablation studies are conducted to demonstrate the efficacy of this procedure, as detailed in Section \ref{sec:ab} and Appendix \ref{sec:moreab}.
The training and inference procedures are outlined in Algorithm \ref{alg:QT}, providing a comprehensive summary of the processes involved.

\section{Experiment}

In this section, we present an extensive evaluation of our proposed QT model using the widely recognized D4RL benchmark \citep{fu2020d4rl}. 
Our main objective is to assess the effectiveness of QT across various domains, setting it against two prevalent algorithms:  Q-learning methods and CSM algorithms.
Each of these algorithms demonstrates proficiency in specific domains while exhibiting sub-optimal performance in others.
Additionally, we execute an empirical ablation study to dissect and understand the individual contributions of the core components of our methodology.

\begin{table*}[!t]
\renewcommand{\arraystretch}{1.0}
  \centering
  \caption{Ablation on the role of different components. Average and standard deviation scores are reported over 3 seeds for the walker2d-medium-replay task. 'CTP' refers to the Conditional Transformer Policy as detailed in Section \ref{sec:CTP}, 'none' indicates the absence of the Q-value module in the configuration, and 'Inf.' is short for inference.}
   \vspace{0.1cm}
  \label{tab:component}
  \scalebox{0.9}{
  \begin{tabular}{cccccc}
    \hline
      Exp & Policy & Q-value Update & Train with Q-value & Inf. with Q-value & Performance \\
    \hline
     1 & BC  & none & & & 32.3 $\pm$ 9.8 \\
     2 & BC  & n-step & \ding{51} & \ding{51} & 82.2 $\pm$ 0.5 \\ 
     3 & CTP & none & & &  79.4 $\pm$ 2.0 \\
     4 & CTP & n-step & & \ding{51} & 87.6 $\pm$ 1.1 \\
     5 & CTP & n-step & \ding{51} &  &  97.7 $\pm$ 0.3 \\
     6 & CTP & 1-step & \ding{51} & \ding{51} & 85.6 $\pm$ 1.7 \\
     7 & CTP & n-step & \ding{51} & \ding{51} & 98.5 $\pm$ 1.1 \\
    \hline
  \end{tabular}}
  \vspace{-0.2cm}
\end{table*}

\textbf{Datasets.}
We consider five different domains of tasks in D4RL benchmark: Gym, Adroit, Kitchen, Maze2D, and AntMaze. 
The Gym-MuJoCo locomotion tasks, commonly used as standard benchmarks, are relatively straightforward and characterized by datasets with a significant proportion of near-optimal trajectories and smooth reward functions. 
In contrast, the Adroit datasets, primarily derived from human behaviors, exhibit a limited state-action space, necessitating robust policy regularization to maintain agent performance within the expected range.
The Kitchen environment poses a multi-task challenge, requiring the agent to complete four sequential sub-tasks to achieve a desired state configuration, thereby emphasizing the importance of generalization to unseen states rather than relying purely on trajectories seen during training.
Maze2D tasks are designed to assess an offline RL algorithm’s capability to effectively stitch together sub-trajectories to identify the shortest path to a set goal. 
Lastly, AntMaze presents more demanding scenarios with sparse rewards, substituting the simpler 2D ball in Maze2D for a complex 8-DoF ``Ant" quadruped robot, thereby elevating the difficulty level.

\begin{table*}[htp]
  \centering
  \caption{Ablation on the stitching ability. Average and standard deviation scores are reported over 3 seeds for the Maze2D tasks. This encompasses four increasingly complex mazes—open, umaze, medium, and large—each with two reward functions: normal and dense. The highest average scores are highlighted in bold.}
   \vspace{0.1cm}
  \label{tab:stitch}
  \scalebox{0.9}{
  \begin{tabular}{clcp{0mm}cp{0mm}cp{0mm}c}
    \hline
       & Dataset & CQL & & DT & & QDT && QT \\
    \hline
    \multirow{4}*{\RotText{Sparse Reward}}
        & maze2d-open-v0        & $216.7\pm80.7$ && $196.4\pm39.6$ && $190.1\pm37.8$ && \textbf{497.9} $\pm$ 12.3 \rule[0mm]{0mm}{3.mm}\\ 
        & maze2d-umaze-v1       & $94.7\pm23.1$  && $31.0\pm21.3$  && $57.3\pm8.2$ && \textbf{105.4} $\pm$ 4.8 \rule[0mm]{0mm}{3.mm}\\
        & maze2d-medium-v1      & $41.8\pm13.6$  && $8.2\pm4.4$   && $13.3\pm5.6$ && \textbf{172.0} $\pm$ 6.2 \rule[0mm]{0mm}{3.mm}\\
        & maze2d-large-v1       & $49.6\pm8.4$   && $2.3\pm0.9$    && $31.0\pm19.8$ && \textbf{240.1} $\pm$ 2.5 \rule[0mm]{0mm}{3.mm}\\
    \hline
    \multirow{4}*{\RotText{Dense Reward}}
        & maze2d-open-dense-v0  & $307.6\pm43.5$ && $346.2\pm14.3$ && $325.7\pm61.4$ && \textbf{608.4} $\pm$ 1.9 \rule[0mm]{0mm}{3.mm}\\
        & maze2d-umaze-dense-v1 & $72.7\pm10.1$  && $-6.8\pm10.9$  && $58.6\pm3.3$ && \textbf{103.1} $\pm$ 7.8 \rule[0mm]{0mm}{3.mm}\\
        & maze2d-medium-dense-v1& $70.9\pm9.2$   && $31.5\pm3.7$   && $42.3\pm7.1$ && \textbf{111.9} $\pm$ 1.9 \rule[0mm]{0mm}{3.mm}\\
        & maze2d-large-dense-v1 & $90.9\pm19.4$  && $45.3\pm11.2$  && $62.2\pm9.9$ && \textbf{177.2} $\pm$ 7.8 \rule[0mm]{0mm}{3.mm}\\
    \hline
  \end{tabular}}
\end{table*}

\textbf{Baselines.}
We benchmark against a diverse array of baseline methods, each excelling in specific domain tasks.
For policy regularization-based approaches, our selection includes IQL \citep{IQL}, BCQ \citep{BCQ}, BEAR \citep{BEAR}, TD3+BC \citep{TD3BC}, and O-RL \citep{onesteprl}. 
We also consider the CQL \citep{CQL} for Q-value constraint methods.
In the realm of model-based offline RL, we evaluate against MoRel \citep{MoRel} and COMBO \citep{Combo}.
For CSM approaches, our comparisons include DT \citep{DT}, StAR \citep{StAR}, QDT \citep{QDT}, GDT \citep{GDT}, and CGDT \citep{wang2023critic}. 
Additionally, we assess diffusion-based methods such as Diffuser \citep{Diffuser}, DD \citep{DD}, and Diffusion-QL \citep{DQL}.
The performance scores for these baseline methods are sourced either from the best results published in their respective papers or from our own runs, ensuring a fair comparison.

\subsection{Main Results}

We compare our QT with the baselines on five domains of tasks and report the results in Table \ref{tab:main_result}. 
To ensure fair comparisons, we normalize the scores according to the protocol established in \citet{fu2020d4rl}, where a score of 100 corresponds to an expert policy. 
We give the analysis based on each specific domain.

\textbf{Results for Gym Domain.}
We can see while most baseline models demonstrate proficiency on Gym tasks, QT often achieves further enhancements, particularly in `medium' and `medium-replay' tasks, surpassing other Transformer-based methods by a large margin.
It's noteworthy that these datasets encompass trajectories generated by an online SAC \citep{SAC} agent, trained to reach roughly one-third of an expert's performance.
Consequently, other Transformer-based methods typically underperform compared to Q-learning approaches in the absence of an ample quantity of high-quality trajectories \citep{emmons2021rvs}, as seen in the medium-expert dataset.
As elucidated in Section \ref{sec:method}, the incorporation of a policy improvement term in QT directs the policy towards optimal actions within the explored action space subset, significantly contributing to QT's commendable empirical performance.

\textbf{Results for Adroit and Kitchen Domain.}
In the Adroit domain, where offline RL is particularly challenged by extrapolation error due to the limited scope of human demonstrations \citep{fu2020d4rl}, robust policy regularization is essential. 
Our Transformer-based policy, employing the DT loss $\Ls_{DT}$, significantly outperforms diffusion-based baselines. 
This superiority is attributable to its high expressiveness and more effective policy regularization. 
Furthermore, the Kitchen tasks, which demand generalization to unseen states and long-term value optimization, also witness notable performance improvements with QT, underscoring its adaptability and effectiveness in this domain.

\textbf{Results for Maze2D and AntMaze Domain.}
The Maze2D domain serves as a benchmark to evaluate the capacity of offline RL algorithms to effectively stitch segments of disparate trajectories \citep{fu2020d4rl}. 
Integrating the Q-value module with the Transformer policy enhances its ability to navigate the shortest path to the goal using pre-collected sub-trajectories. 
The AntMaze domain, characterized by sparse rewards and an abundance of sub-optimal trajectories, presents a more difficult challenge.
A robust and stable Q-learning approach is essential for achieving notable performance in this setting.
Empirically, QT, augmented with our Q-value module and an optimally tuned hyper-parameter $\eta$, either matches or exceeds the performance of existing methods, whereas other Transformer-based approaches often struggle in `medium' and `large' tasks.

\begin{table*}[h]
    \small
    \caption{Ablation on the sparse reward ability. 
    Average and standard deviation scores are reported over 3 seeds for the D4RL tasks.
    The study includes three tasks—halfcheetah, hopper, and walker2d—each evaluated under two reward conditions: sparse and dense. 
    The highest average scores are denoted in bold.
    }
     \vspace{0.1cm}
    \label{tab:sparse_reward}
    \centering
    \scalebox{0.88}{
    \begin{tabular}{c|cccc|cccc}
    \toprule
    \multirow{2}{*}{\textbf{Dataset}} &
    \multicolumn{4}{c|}{\textbf{Sparse Reward}}  &
    \multicolumn{4}{c}{\textbf{Dense Reward}}  \\ 
    & {DT} & {CQL} & {QDT} & {QT} & {DT} & {CQL} & {QDT} & {QT}  \\
    \midrule
    halfcheetah-medium-v2 & 42.2 $\pm$ 0.2 & 1.0 $\pm$ 1.0 & 42.4 $\pm$ 0.5 & \textbf{43.3 $\pm$ 0.2} & 42.6 $\pm$ 0.1 & 49.2 $\pm$ 0.5 & 42.3 $\pm$ 0.4 & \textbf{51.4 $\pm$ 0.4} \\
    hopper-medium-v2 & 57.3 $\pm$ 2.4 & 23.3 $\pm$ 1.0 & 50.7 $\pm$ 5.0 & \textbf{72.7 $\pm$ 3.9} & 67.6 $\pm$ 1.0 & 69.4 $\pm$ 13.1 & 66.5 $\pm$ 6.3 & \textbf{96.3 $\pm$ 3.1} \\
    walker2d-medium-v2 & 69.9 $\pm$ 2.0 & 0.0 $\pm$ 0.4 & 63.7 $\pm$ 6.4 & \textbf{80.7 $\pm$ 0.8} & 74.0 $\pm$ 1.4 & 83.0 $\pm$ 0.6 & 67.1 $\pm$ 3.2 & \textbf{88.8 $\pm$ 0.5} \\ \midrule
    halfcheetah-medium-replay-v2 & 33.0 $\pm$ 4.8 & 7.8 $\pm$ 6.9 & 32.8 $\pm$ 7.3 & \textbf{42.5 $\pm$ 0.2} & 36.6 $\pm$ 0.8 & 45.5 $\pm$ 0.5 & 35.6 $\pm$ 0.5 & \textbf{48.9 $\pm$ 0.3} \\
    hopper-medium-replay-v2 & 50.8 $\pm$ 14.3 & 7.7 $\pm$ 5.9 & 38.7 $\pm$ 26.7 & \textbf{94.2 $\pm$ 2.2} & 82.7 $\pm$ 7.0 & 95.0 $\pm$ 2.9 & 52.1 $\pm$ 20.3 & \textbf{102.0 $\pm$ 0.2} \\
    walker2d-medium-replay-v2 & 51.6 $\pm$ 24.6 & 3.2 $\pm$ 1.7 & 29.6 $\pm$ 15.5 & \textbf{78.5 $\pm$ 2.1} & 66.6 $\pm$ 3.0 & 77.2 $\pm$ 1.1 & 58.2 $\pm$ 5.1 & \textbf{98.5 $\pm$ 1.1} \\
    \midrule
    Average & 50.8 & 7.2 & 43.0 & \textbf{68.6} & 61.7 & 69.9 & 53.6 & \textbf{81.0}  \\
    \bottomrule
    \end{tabular}}
\end{table*}

\subsection{Ablation Study}
\label{sec:ab}

This section delves into a quantitative analysis of QT's superior performance over other Transformer-based methods on D4RL tasks. 
We undertake an ablation study to dissect and quantify the contributions of QT's main components to its overall efficacy. 
Additionally, further ablations are conducted to assess whether QT successfully integrates the strengths of both CSM and Q-learning methods while overcoming their limitations.
We select CQL as the benchmark for evaluating the Q-learning approach, and DT as the benchmark for assessing the CSM approach.
We also include QDT as a comparative benchmark in order to showcase the differences between QDT and our approach.
Note that further discussion about QT is provided in Appendix \ref{sec:moreab}.

\textbf{Role of Different Components.}
As delineated in Section \ref{sec:method}, our methodology comprises three primary components, alongside the Q-value update method, each warranting individual analysis. 
We select the walker2d-medium-replay dataset as the benchmark due to its diverse range of agent levels and the substantial performance enhancement QT demonstrates compared to baselines. 
As indicated in Table \ref{tab:component}, integrating our Q-value module significantly boosts performance, as evidenced by the comparative results between experiments 1 vs. 2, and 3 vs. 7.
Notably, the Q-value regularization (Equation \ref{eq:final_update}) during the training stage is instrumental, manifesting as the most significant contributor to performance enhancement, with the inference phase also benefiting from the Q-value module (as seen in comparisons among experiments 3 vs. 4, and 5 vs. 7).
Furthermore, relying solely on the 1-step Bellman equation for updating the Q-value function results in subpar performance compared to the n-step Bellman equation (as seen in comparisons between experiments 6 and 7), which underscores the criticality of Q-value function accuracy in our methodology.

\textbf{Stitching Ability.}
The Maze2D domain, a navigation task with a fixed goal location, serves as a critical test for offline RL algorithms' ability to stitch together different trajectory segments \citep{fu2020d4rl}. 
This domain comprises four increasingly complex mazes—open, umaze, medium, and large—and utilizes two reward functions: normal and dense. 
The normal reward is granted solely upon goal achievement, while the dense reward is incrementally distributed at each step, inversely proportional to the distance from the goal.
Table \ref{tab:stitch} summarizes the results. 
CQL performs notably well, particularly with dense rewards. 
DT, however, often struggles due to its limited stitching capability. 
QDT demonstrates a marked improvement over DT but still lags behind CQL. 
Significantly, QT excels across all tasks, affirming its ability to not only endow the Transformer policy with stitching capacity but also synergistically merge the strengths of both methodologies for enhanced performance.

\textbf{Sparse Reward Ability.}
To illustrate the limitations of the Q-learning approach (CQL), we follow \citet{DT} and evaluate the algorithms in a delayed (sparse) reward setting, where rewards are withheld during the trajectory and aggregated at the final timestep.
Table \ref{tab:sparse_reward} presents the results for both delayed (sparse) and dense reward scenarios. 
As anticipated, CQL exhibits difficulty in formulating an effective policy under sparse conditions, in contrast to DT, which demonstrates commendable performance. 
QDT, which employs CQL for RTG token value relabeling, registers inferior performance compared to DT, influenced by CQL's inaccurate value function estimations. 
Conversely, QT, while similarly impacted by these inaccurate estimations in sparse reward scenarios, benefits from our robust policy regularization. 
This feature effectively mitigates the adverse effects of the Q-value module, enabling QT to outperform these methods across all assessed tasks.

\begin{figure}[ht!]
    \centering
    \includegraphics[width=0.95\linewidth]{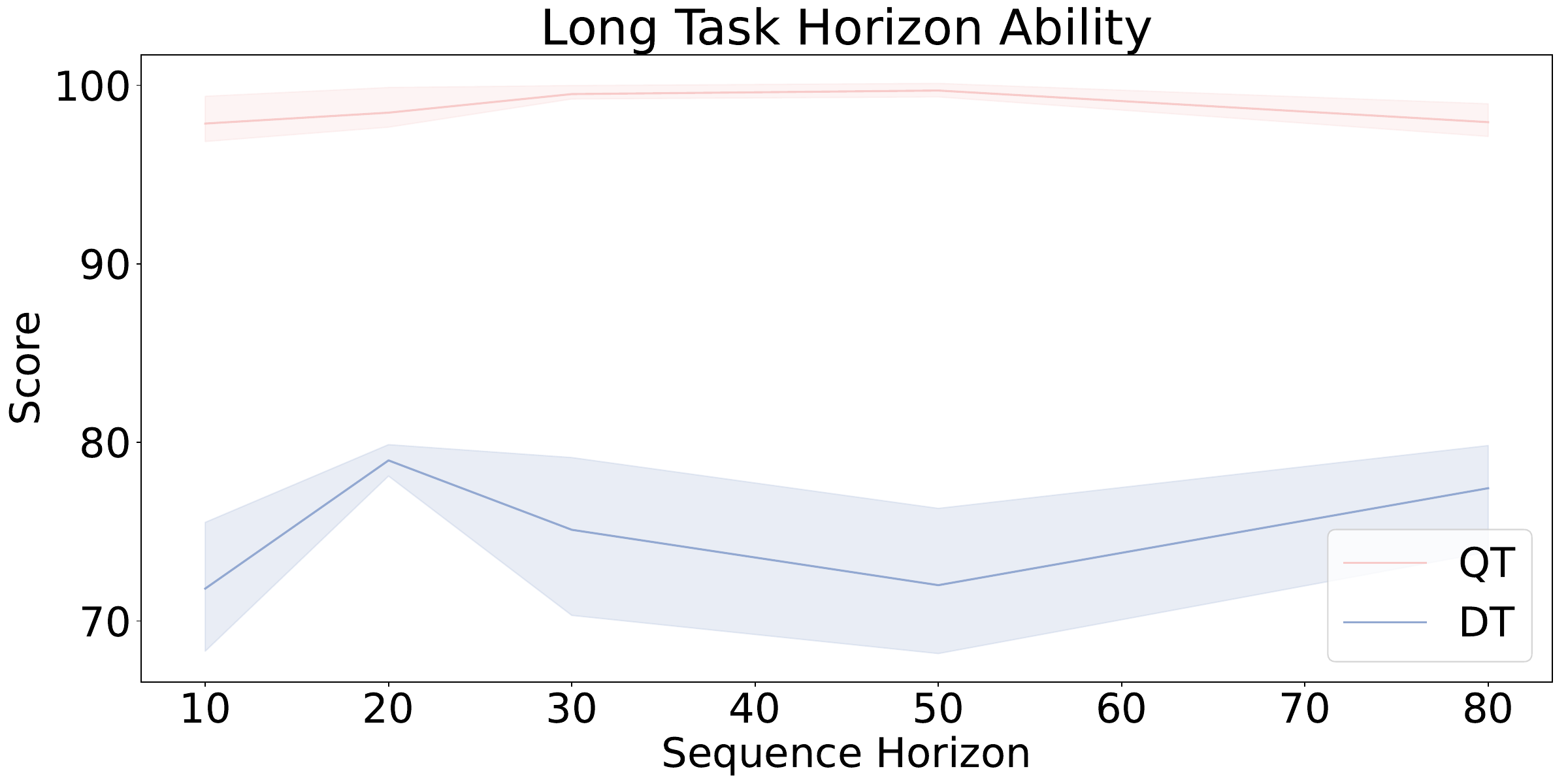}
    \vspace{-0.3cm}
    \caption{Ablation on the long task horizon ability. This encompasses the performance comparison of different input sequence horizons $K \in [10, 80]$ in the walker2d-medium-replay-v2 task.}
    \label{fig:length}
    \vspace{-0.4cm}
\end{figure}

\textbf{Long Task Horizon Ability.}
While in a Markovian environment, the state at the previous moment is often sufficient to determine the current action, the DT experiment reveals that past information is valuable for the sequence modeling method in some environments, where longer sequences tend to yield better results than those of length 1.
We then explore the impact of different sequence lengths on performance and compare the results of DT and QT, where Q-learning methods often perform badly in the long horizon setting \citep{QDT, bhargava2023sequence}.
The results are shown in Figure \ref{fig:length}.
As the sequence horizon $K$ extends, both agents exhibit improved performance. 
DT initially deteriorates after $K=20$ but recovers at $K=80$, whereas QT consistently enhances its performance, demonstrating a superior capability to manage extended task horizons.
\section{Related Work}

Offline RL algorithms learn a policy entirely from this static offline dataset $\gD$, without online interactions with environment \citep{levine2020offline}. 
This paradigm can be precious in case the interaction with environment is expensive or high-risk (e.g., safety-critical applications).
However as the learned policy might differ from the behavior policy, the offline algorithms must mitigate the effect of the \textit{distribution shift}, which can result in a significant performance drop, as demonstrated in prior research \citep{fujimoto2019off}.

\textbf{Q-learning} method is one of the most prominent categories to address the \textit{distribution shift} problem. 
Especially, previous Q-learning works generally address this problem in one of three ways:
1) constraining the learned policy to the behavior policy \citep{BEAR, fujimoto2019off, TD3BC, wu2019behavior, lyu2022mildly};
2) constraining the learned policy by making conservative estimates of future rewards \citep{CQL, kostrikov2021offline, qtransformer};
3) introducing model-based methods, which learn a model of the environment dynamics to generate more data for policy training 
and perform pessimistic planning in the learned MDP \citep{janner2019trust, MoRel, Combo}.

\textbf{Weighted imitation learning} addresses the \textit{distribution shift} without restricting the learned policy, which carries out imitation learning by putting higher weights on the good state-action pairs.
These methods \citep{wang2018exponentially, peng2019advantage, wang2020critic, chen2020bail, siegel2020keep} usually use an estimated advantage function as the weight. 
As these approaches imitate the selected parts of the behavior policy, they naturally restrict the learned policy within the behavior policy.

\textbf{Conditional sequence modeling} is the other group of approaches without restricting the learning policy, which predicts subsequent actions from a sequence of past experiences, encompassing state-action-reward triplets.
This paradigm lends itself to a supervised learning approach, inherently constraining the learned policy within the boundaries of the behavior policy and focusing on a policy conditioned on specific metrics for future trajectories \citep{DT, PTDT, RCSL, CommFormer, meng2023offline, wang2023critic}.
Moreover, the sequence of trajectories could also be formulated as a conditional generative process and generated by the diffusion model while satisfying conditioned constraints \citep{Diffuser, DD, DQL}.

Our approach is distinct from but related to these primary classes of offline RL algorithms.
Essentially, our method is a CSM approach as it learns the subsequent actions based on historical sequences and sampled future rewards.
Also, the high-level framework of our approach is somewhat akin to weighted imitation learning, wherein a value function is employed to assign weights to various state-action pairs.
However, the practical application of our components markedly differs.
Unlike approaches that use the value function merely for training data weighting, our method integrates a learned Q-value module directly into the training phase, which biases action sampling towards higher-return options, a factor that has empirically demonstrated enhanced performance in our experiments.

\section{Conclusion}

In this study, we introduce QT, which combines the trajectory modeling ability of Transformer with the predictability of optimal future returns from DP methods. 
QT offers a novel framework for enhancing offline RL algorithms. 
The Conditional Transformer Policy of QT allows for a highly expressive policy class whose learning itself acts as a strong policy regularization method.
Additionally, the integration of a Q-value regularization via a jointly learned Q-value function biases action sampling towards optimal regions within the exploration space. 
Empirical evaluations on D4RL benchmark datasets demonstrate the superiority of QT over traditional DP and CSM methods, highlighting the potential of QT to enhance the SOTA in offline RL.

\textbf{Limitation. ~~} We introduce a novel Transformer-based policy for offline RL, achieving state-of-the-art performance across various tasks. However, QT's efficacy depends on the availability of explicit reward signals. In scenarios lacking explicit reward signals, such as datasets containing only state-action pairs from human demonstrations, QT's performance may be limited.

\section*{Acknowledgements}
This work is supported by the National Key R\&D Program of China (No. 2022ZD0160702), STCSM (No. 22511106101, No. 22511105700, No. 21DZ1100100), 111 plan (No. BP0719010) and National Natural Science Foundation of China (No. 62306178).

\section*{Impact Statement}
This paper contributes to the advancement of Offline Reinforcement Learning.
While there are many potential societal consequences of our work, we believe that none require specific emphasis in this context.

\bibliography{example_paper}
\bibliographystyle{icml2024}

\newpage
\appendix
\onecolumn

\section{Proofs}
\label{sec:proof}

\subsection{Proof of Theorem \ref{thm:onlydt}}

First we introduce the following Lemma, which is motivated by the work of \citet{RCSL} on return-conditioned supervised learning (RCSL).
\begin{lemma}\citep{RCSL}
\label{lem:infinite}
Consider an MDP, behavior $ \beta$, and conditioning function $ f$. Assume the following:
\begin{enumerate}
    \item Return coverage: $g(\tau) = \sum_{t=1}^\gH r_t$, $ P_\beta(g=f(\rvs_1)|\rvs_1) \geq \alpha_f$ for all initial states $ \rvs_1$.
    \item Near determinism: $ P(r \neq \gR(\rvs, \rva) \text{ or } \rvs' \neq \gT(\rvs, \rva) | \rvs, \rva ) \leq \epsilon$ at all $ \rvs, \rva $ for some functions $ \gR $ and $ \gT $. Note that this does not constrain the stochasticity of the initial state.
    \item Consistency of $ f$: $ f(\rvs) = f(\rvs') + r$ for all $ \rvs$.\footnote{Note this can be exactly enforced (as in prior work) by augmenting the state space to include the cumulative reward observed so far.}
\end{enumerate}
Let $J(\pi) = \E_{\tau \sim \pi} [g(\tau)]$, then
\begin{align}
\label{eq:lemma1}
    \E_{\rvs_1}[f(\rvs_1)] - J(\pi_f^{RCSL}) \leq \epsilon\left( \frac{1}{\alpha_f} + 2\right)\gH^2.
\end{align}
\end{lemma}

Using the above lemma, we can prove the Theorem \ref{thm:onlydt}.

\begin{proof}

Considering the offline dataset collected by the behavior policy $\beta$, we choose the condition function $f$ as $f(\rvs_1) = \sum_{r_{1:\gH} \sim \pi_\beta (\rvs_1)} r$ and plug it into the left part of Equation \ref{eq:lemma1}, we can see:
\begin{align}
    \E_{\rvs_1}[f(\rvs_1)] - J(\pi_f^{RCSL}) &= \E_{\rvs_1}[\sum_{r_{1:\gH} \sim \pi_\beta (\rvs_1)} r] - J(\pi_f^{RCSL}) \\
    &= \E_{\tau \sim \pi_\beta} [\sum_{t=1}^{\gH} r_t] - J(\pi_f^{RCSL}) \\
    &= \E_{\tau \sim \pi_\beta} [g(\tau)] - J(\pi_f^{RCSL})
\end{align}
Then consider the reward-to-go $\hat{r}_t = \sum_{i=1}^\gH r_i$ defined in the Equation \ref{eq:input}, it is obvious that the condition function  $\hat{r}_t$ satisfies the requirement about the consistence of conditioning function, which we could get the following Equation:
\begin{align}
    \E_{\rvs_1}[f(\rvs_1)] - J(\pi_f^{RCSL}) = \E_{\tau \sim \pi_\beta} [g(\tau)] - \E_{\tau \sim \pi_f} [g(\tau)] \leq \epsilon (\frac{1}{\alpha_f} + 2) \gH^2
\end{align}
Combining this with Lemma \ref{lem:infinite} yields the result.

\end{proof}

\subsection{Proof of Theorem \ref{thm:qdt}}

Motivated by the proof in \citet{hu2023iteratively}, we first give some lemmas to help the proof of Theorem \ref{thm:qdt}.

We consider a $k$-armed one-step decision-making problem.
Let $\Delta$ be a $k$-dimensional simplex and $\vq=(q(1),\dots,q(k)) \in\sR^k$ be the reward vector. 
The final optimization considers:
\begin{align}
     \max_{\pi \in \Delta} \pi \cdot \vq + \tau \sH(\pi).
\end{align}

The next result characterizes the solution of this problem (Lemma 4 of \citet{nachum2017bridging}).

\begin{lemma}\citep{nachum2017bridging}
\label{lem:nachum-softmax}
For $\tau > 0$, 
let  
\begin{align}
F_{\tau}(\vq) = \tau \log  \sum_{a} e^{ q(a) / \tau} \, ,
\quad
f_{ \tau}(\vq) = \frac{e^{\vq / \tau}}{\sum_{a} e^{q(a) / \tau}} = e^{\frac{\vq - F_{\tau}(\vq)}{\tau}}\, .
\end{align}
Then there is
\begin{align}
F_{ \tau}(\vq) = \max_{\pi\in\Delta}\ \pi\cdot \vq + \tau \sH(\pi)
=
f_{\tau}(\vq)\cdot \vq + \tau \sH(f_{\tau}(\vq))\, .
\end{align}
\end{lemma}

The second result provides the error decomposition when applying the Politex algorithm to compute an optimal policy, as adopted from \citet{abbasi2019politex}.

\begin{lemma} \citep{hu2023iteratively}
Let $\pi_0$ be the uniform policy and consider running the following iterative algorithm on a MDP for $t\geq 0$,
\begin{align}
\pi_{t+1}(\rva | \rvs) \propto  \pi_{t} (\rva|\rvs) \exp\left(  \frac{{q}^{\pi_t} (\rva|\rvs) }{\tau} \right)\, ,
\end{align}
Then
\label{lem:politex}
\begin{align}
v^*(\rvs) - v^{\pi_t}(\rvs) \leq \frac{1}{(1-\gamma)^2} \sqrt{\frac{2\log |\gA|}{t}} \, .
\end{align}
\end{lemma}

Using the above lemmas, we can prove the Theorem \ref{thm:qdt}.

\begin{proof}

First recall the in-sample optimality equation
\begin{align}
q_{\pi_\beta}^{*}(\rvs,\rva) = \gR (\rvs,\rva) + \gamma \E_{\rvs' \sim \gT(\cdot|\rvs,\rva)} \left[ \max_{\rva': \pi_\beta(\rva'|\rvs') > 0} q_{\pi_\beta}^{*}(\rvs', \rva') \right], 
\end{align}
which could be viewed as the optimal value of a MDP $M_{\gD}$ covered by the behavior policy $\pi_{\beta}$, where $M_{\gD}$ only contains transitions starting with $(\rvs,\rva) \in \gS \times \gA$ such that $\pi_{\beta}(\rva|\rvs) > 0$.  
Then the result can be proved by two steps. 
First, the QT algorithm will never consider actions such that $\pi_{\beta}(\rva|\rvs) = 0$. 
This is directly implied by \cref{lem:nachum-softmax}. 
Second, we apply \cref{lem:politex} to show the error bound of using QT on $M_{\gD}$, which implies that $V^{\pi^*} (\rvs) \geq V^{\beta} (\rvs)$.
This finishes the proof. 

\end{proof}

\section{Implementation Details}

\textbf{Conditional Transformer Policy.} We build our policy as a Transformer-based model, which is based on minGPT open-source code \footnote{\url{https://github.com/karpathy/minGPT}}.
The detailed model parameters are in Table \ref{tab:model_parameter}.

\textbf{Q networks.} We build two Q networks with the same MLP setting as our diffusion policy, which has 3-layer MLPs with Mish activations and 256 hidden units for all networks.

We use the Adam \citep{Adam} optimizer for the training of both Conditional Transformer Policy and Q networks.

\begin{table}[h]
\renewcommand{\arraystretch}{1.3}
\centering
  \caption{Hyperparameters of QT in our experiments.}
  \vspace{0.1cm}
  \label{tab:model_parameter}
  \begin{tabular}{ll}
    \hline
    Parameter & Value \\
    \hline
    Number of layers            & 4 \\
    Number of attention heads   & 4 \\
    Embedding dimension         & 256 \\
    Nonlinearity function       & ReLU \\
    Batch size                  & 256 \\
    Context length $K$          & 20 \\
    Dropout                     & 0.1 \\
    Learning rate               & 3.0e-4 \\
    \hline
  \end{tabular}
\end{table}

\section{Hyper-parameters}

For QT, we consider two hyper-parameters in total: Q-value regularization weight $\eta$ and gradient normalization. For the Q-value regularization weight $\eta$, we consider values according to the characteristics of different domains, and we also conduct simple ablations to investigate how to choose the value. 
As indicated in  Equation \ref{eq:final_update}, $\eta$ is a critical hyper-parameter that balances policy regularization and policy improvement losses.
The walker2d-medium-replay dataset, in both dense and sparse reward scenarios, is selected for benchmarking. 
Table \ref{tab:eta} displays the outcomes, illustrating QT's sensitivity to $\eta$ selection, with varying values yielding significantly different performances. 
A larger $\eta$ enhances performance when the Q-value is accurately estimated within the dataset. 
Conversely, in scenarios like sparse rewards where Q-value estimation is challenging, a smaller $\eta$ proves more efficacious.
For the gradient normalization, we consider values in the grid $\{5.0, 9.0, 15.0, 20.0 \}$.
Based on these considerations, we provide our hyper-parameter setting in Table \ref{tab:hyperparameter}.

\begin{table}[h]
  \caption{Ablation on the role of the hyper-parameter $\eta$. Average and standard deviation scores are reported over 3 seeds for the walker2d-medium-replay task.}
  \vspace{0.1cm}
  \label{tab:eta}
  \centering
  \scalebox{1.0}{
  \begin{tabular}{lccccc}
    \hline
     $\eta$ & 0.01 & 0.1 & 1 & 2 & 3 \\
    \hline
    dense & $88.0 \pm 0.4$ & $89.2 \pm 1.0$ & $95.4 \pm 0.5$ & $98.5 \pm 1.1$ & $98.4 \pm 0.4$  \\
    sparse & $78.5 \pm 2.1$ & $72.3 \pm 0.3$ & $7.0 \pm 4.6$ & $8.5 \pm 2.5$ & $10.6 \pm 6.1$   \\
    \hline
  \end{tabular}}
\end{table}

\begin{table}[!h]
    \caption{\small Hyperparameter settings of all selected tasks. }
    \vspace{0.1cm}
    \label{tab:hyperparameter}
    \centering
    \resizebox{0.8\textwidth}{!}{
    \begin{tblr}{
    colspec = {l|*{2}{c}|l|*{2}{c}},
    row{1} = {font=\bfseries}
    }
    \toprule
        Tasks & $\eta$ & grad norm & Tasks & $\eta$ & grad norm   \\
        \hline
        halfcheetah-medium-expert-v2 & 2.5 & 15.0 & pen-human-v1 & 0.1 & 9.0  \\ 
        hopper-medium-expert-v2 & 1.0 & 9.0  & hammer-human-v1 & 0.1 & 5.0   \\ 
        walker2d-medium-expert-v2 & 2.0 & 5.0 & door-human-v1 & 0.005 & 9.0   \\ 
        
        halfcheetah-medium-v2 & 5.0 & 15.0 & pen-cloned-v1 & 0.1 & 9.0 \\
        hopper-medium-v2 & 1.0 & 9.0 & hammer-cloned-v1 &  0.01 & 9.0 \\ 
        walker2d-medium-v2 & 2.0 & 5.0 & door-cloned-v1 & 0.001 & 9.0 \\
        
        halfcheetah-medium-replay-v2 & 5.0 & 15.0 & kitchen-complete-v0 & 0.005 & 9.0 \\
        hopper-medium-replay-v2 & 3.0 & 9.0 & kitchen-partial-v0 & 0.01 & 9.0 \\
        walker2d-medium-replay-v2 & 2.0 & 5.0 & - & - & - \\

        maze2d-open-v0 & 0.01 & 9.0  & maze2d-open-dense-v0 & 0.01 & 9.0  \\
        maze2d-umaze-v1 & 5.0 & 20.0 & maze2d-umaze-dense-v1 & 5.0 & 20.0   \\
        maze2d-medium-v1 & 5.0 & 9.0 & maze2d-medium-dense-v1 & 5.0 & 9.0   \\
        maze2d-large-v1 & 4.0 & 9.0 & maze2d-large-dense-v1 & 4.0 & 9.0   \\

        antmaze-umaze-v0 & 0.05 & 9.0  & antmaze-medium-diverse-v0 & 0.01 & 9.0  \\
        antmaze-umaze-diverse-v0 & 0.01 & 9.0 & antmaze-large-diverse-v0 & 0.005 & 9.0    \\
    \bottomrule
    \end{tblr}}
\end{table}

\section{Further Discussions}
\label{sec:moreab}

\subsection{Performance of QT in the Atari Environment}

Recognizing the importance of discrete action domains in RL, we expand our investigation to include Atari games, a domain characterized by its high-dimensional visual inputs and the delayed reward challenge. 
We benchmark our QT method against established baselines that are evaluated in the DT method, normalizing scores where 100 represents a professional gamer's score and 0 denotes a random policy. 
As detailed in Table \ref{tab:atari}, our findings demonstrate that QT consistently achieves competitive performance, affirming its efficacy in discrete action domains.

\begin{table*}[]
\centering
\caption{Results for 1\% DQN-replay Atari datasets. We evaluate the performance of QT on four Atari games using three different seeds, and report the mean and variance of the results. 
The best mean scores are highlighted in bold.}
\label{tab:atari}
\vspace{.1cm}
\begin{tabular}{lcccccc}
\toprule
Game     & CQL   & QR-DQN & REM & BC           & DT           & \textbf{QT}   \\ \midrule
Breakout & 211.1 & 17.1   & 8.9 & 138.9 $\pm$ 61.7  & 267.5 $\pm$ 97.5 &    \textbf{423.9 $\pm$ 87.2}   \\
Qbert    & \textbf{104.2}& 0      & 0   & 17.3 $\pm$ 14.7  & 15.4 $\pm$ 11.4   &    46.7 $\pm$ 13.3     \\
Pong     & \textbf{111.9} & 18     & 0.5 & 85.2 $\pm$ 20.0  & 106.1 $\pm$ 8.1  &  108.3 $\pm$ 2.0              \\
Seaquest & 1.7   & 0.4    & 0.7 & 2.1 $\pm$ 0.3    & 2.5 $\pm$ 0.4    & \textbf{  4.0 $\pm$ 0.3 }   \\ 
\midrule
\textbf{Average} & 107.2 & 8.9 & 2.5 & 69.9 & 97.9 & \textbf{145.7}  \\ \bottomrule
\end{tabular}%
\end{table*}

\subsection{Conditional Action Generation}

In pure DT approaches, the generation of diverse actions is conditioned on varying RTG values due to its trajectory-level modeling. While this method offers diversity, it faces the challenge of unmatched RTG values, requiring significant human effort to identify the optimal RTG for each scenario. Our QT method strategically avoids the manual selection of RTG values, which often relies heavily on prior knowledge and can be labor-intensive, streamlining the learning process and reducing dependency on manual intervention.

Specifically, QT addresses these challenges by integrating a Q-value maximization step within the training phase, guiding the CSM policy toward generating actions aligned with optimal return objectives. Just as Table \ref{tab:differrtg} shows, this adjustment enhances the policy's efficacy and reduces the reliance on precise RTG selection within a certain range, providing a more efficient approach to action generation.
However, QT may still encounter difficulties when there is a significant deviation between the selected RTG and the optimal trajectory. Despite this, the QT framework incorporates a Q-value function during the inference stage, offering a dynamic and adaptive strategy to ascertain optimal actions, thus augmenting the method's practicality and reducing the need for extensive manual calibration.

\begin{table*}[]
\centering
\caption{Ablation on the conditional action generation. Average and standard deviation scores are reported over 3 seeds for the walker2d-medium-replay task. QT* indicates that only Q-value regularization is included in the training stage.}
\label{tab:differrtg}
\vspace{.1cm}
\begin{tabular}{lcccccc}
\toprule
RTG     & 1000   & 2000 & 3000 & 4000           & 5000          & Infer with Q-value function   \\ \midrule
QT* & 51.0 $\pm$ 1.0 & 68.6 $\pm$ 0.7   & 95.3 $\pm$ 1.1	 & 96.3 $\pm$ 0.4  & 97.2 $\pm$ 0.2 &  98.5 $\pm$ 1.1   \\
DT    & 32.4 $\pm$ 1.2	& 58.8 $\pm$ 0.5  & 75.7 $\pm$ 0.6	 & 79.4 $\pm$ 2.0  & 77.0 $\pm$ 0.6  &    87.6 $\pm$ 1.1     \\
\bottomrule
\end{tabular}%
\end{table*}

\subsection{Differences Between QT and Other Q-Learning Methods}

As delineated in Section \ref{sec:rethink}, QDT \citep{QDT} takes the first attempt to combine the CSM with Q-learning by learning a conservative value function to relable the RTG tokens in the dataset, keeping other components aligned with DT. However, such adaptations essentially constitute simple data augmentation, incorporating ”stitched” trajectories into the training dataset but continuing to encounter unmatched RTG values during inference due to trajectory-level modeling.

Conversely, the Q-Transformer \citep{qtransformer} introduces a nuanced utilization of the transformer architecture to refine the learning of the Q-value function. It achieves this through action discretization, coupled with the novel application of a conservative regularizer. This regularizer is specifically designed to constrain out-of-distribution Q-values, ensuring their proximity to the minimal achievable cumulative rewards. However, the Q-Transformer still remains within the purview of traditional Q-learning methodologies, albeit with a significant enhancement in feature representation capabilities through the adoption of transformer architecture.

For a more granular comparison, Table \ref{tab:difference} elucidates the key distinctions among these methods.

\begin{table*}[]
\centering
\caption{Detailed comparison of QDT, QT, and Q-Transformer.}
\label{tab:difference}
\vspace{.1cm}
\begin{tabular}{L{2.5cm}L{4cm}L{4cm}L{4cm}}
\toprule
Aspect    & QDT   & QT & Q-Transformer \\ \midrule
Training dataset & Augmented with relabeled RTG tokens & Utilizes the original dataset   & Utilizes the original dataset   \\ \midrule
Training loss    & MSE Loss for continuous actions & MSE Loss for continuous actions, supplemented with Q-value function maximization  & TD error coupled with conservative regularization  \\ \midrule
Hindsight info  & Individual Return-to-Go values & A set of candidate Return-to-Go values  & Does not utilize hindsight information  \\ \midrule
Inference & Relies on the transformer's output & Leverages the transformer output with a selection mechanism from the learned Q-value function & Selects from the entire action space through the maximization of the learned Q-value function \\  
\bottomrule
\end{tabular}%
\end{table*}

\subsection{Sparse Reward Setting}

We explore the performance of the QT in environments with varying reward densities, specifically focusing on the maze2d and MuJoCo Gym tasks. Our findings indicate an inconsistency: in sparse settings of maze2d-medium and maze2d-large environments, QT outperforms compared to denser reward configurations, contrary to the observed trend in MuJoCo tasks.

A potential explanation for this discrepancy lies in the fundamental differences between these environments. Maze2d environments, characterized by their simplicity and shorter episode lengths, contrast with the MuJoCo tasks, which feature higher action/state dimensions and longer episode durations, as detailed in Table \ref{tab:env_detail}.

Another potential explanation is the reward structure in the maze2D-dense environments. In these settings, rewards are based on the negative exponentiated distance to the target, potentially inflating the values for `failure' trajectories that approximate the target yet encounter obstacles. Our method, designed to sample high-value actions while adhering to the behavior policy, may inadvertently prioritize these `false' high-value actions, leading to suboptimal performance compared to sparse settings where high-value actions are unequivocally associated with reaching the target. Conversely, in environments like open and umaze, where obstacles are absent, QT demonstrates superior performance in dense settings, supporting this hypothesis.

\begin{table*}[]
\centering
\caption{Comparison of MuJoCo Gym and Maze2D Environments. The table shows the action dimension, state (observation) dimension, and average episode length over the top 5\% returns in the dataset.}
\label{tab:env_detail}
\vspace{.1cm}
\begin{tabular}{lccc}
\toprule
Environment    & Action Dim   & State Dim & Good Episode Average Length \\ \midrule
hopper & 3 &11   & 708.2   \\ 
halfcheetah & 6 & 17 & 1000.0 \\
walker2d    & 6 & 17 & 996.7 \\ \midrule
maze2d-open  & 2 & 4  & 49.8  \\ 
maze2d-umaze  & 2 & 4  & 128.6  \\ 
maze2d-medium  & 2 & 4  & 224.1  \\ 
maze2d-large  & 2 & 4  & 314.6  \\ 
\bottomrule
\end{tabular}%
\end{table*}

\subsection{How QT Improves Stitching Ability}

While our theoretical exposition offers a robust motivation, which posits that the Q-value module serves as a pivotal mechanism for policy improvement, the assertion that QT enhances the stitching capability is primarily evidenced through empirical studies.
In one word, the integration of Q-value regularization with DT addresses the alignment issues inherent in pure CSM approaches to enhance the model's ability to stitch together optimal actions, thereby improving the overall effectiveness and robustness of the policy learned from offline data.

In pure CSM models, the RTG token significantly influences the learning process by providing a trajectory-level perspective. However, this trajectory-centric approach can lead to potential misalignments between the RTG values and the current state-action pairs during inference, potentially leading to suboptimal decision-making \citep{wang2023critic}.
To address this concern and enhance the alignment between learning and inference, we integrate the Q-value function, which offers a granular, state-action specific estimation of future returns. This integration allows for a more dynamic and responsive decision-making process during training and inference, where actions are selected based on their immediate value rather than a predetermined trajectory, and the learning and inference processes are aligned through the learned Q-value function.

During the learning phase, the model is trained to select actions that minimize the combined loss (as outlined in Equation \ref{eq:final_update}), which includes components from both the CSM and Q-value paradigms. This process ensures that the policy is grounded in the distribution of the training dataset while also being attuned to the optimal action values as estimated by the Q-value module.
During inference, the model leverages the learned Q-value function to make decisions. Instead of relying on RTG tokens, the model evaluates a set of candidate actions generated based on various RTG values and selects the one with the highest Q-value. This approach ensures that the decision-making process is informed by both the trajectory modeling insights from the CSM component and the optimal action value estimation from the Q-value component.

Our ablation studies, meticulously documented in Table \ref{tab:component}, provide empirical substantiation for this methodology. When inference relies on the learned Q-value function (Exp 4), it surpasses the performance of a purely CSM-based method (Exp 3), validating the phenomenon of unmatched RTG value in the trajectory-level modeling. Additionally, we have conducted further ablation studies that vary RTG tokens within the context of pure CSM models in Table \ref{tab:differrtg}. These studies are designed to rigorously examine the phenomenon of RTG value misalignment and its impact on the model's performance. Moreover, the integration of the Q-value module throughout both the learning and inference phases aligns the learning objectives with the inference dynamics, which fosters a more robust and effective decision-making framework (Exp. 7 in Table \ref{tab:component}).

\subsection{How QT Addresses Overfitting of the Q-Value Function}

In the training of our Q-value functions, the expected Q-value ($\hat{Q}_m$ in Equation \ref{eq:Q_update}) is derived from the n-step Bellman equation. The action $\rva_t$ is selected according to the target policy $\pi_{\theta'}$, generated by the CSM models. This design ensures that the actions produced by the CSM models predominantly align with the distribution observed in the training dataset (with small $\eta$), thus reducing the risk of overestimating Q-values for out-of-distribution actions. What's more, during the inference, the interplay between the candidate actions generated by the multiple RTG tokens and the Q-value function’s guidance facilitates a more nuanced and effective action selection process, avoiding the pitfalls of direct Q-value maximization.

It is imperative to note that our policy derivation is distinct from traditional Q-learning methodologies. Our policy emerges from the CSM models other than the Q-value function, primarily governed by the MSE loss delineated in Equation \ref{eq:DTLoss}. Here, the Q-value function serves as a component for policy enhancement, with its influence on the final policy modulated by the hyper-parameter $\eta$. In scenarios where data is exceptionally sparse or noisy, which complicates accurate Q-value estimation, modulating $\eta$ can significantly mitigate the adverse effects of overfitting or incorrect Q-value approximation.

To empirically substantiate our claims, we have conducted an ablation study detailed in Table \ref{tab:eta} above. We selected the walker2d-medium-replay dataset in both dense and sparse reward settings. The results demonstrate that in environments conducive to accurate Q-value estimation (dense reward scenario, where the performance of CQL is $77.2$), a higher $\eta$ enhances performance. Conversely, in settings where Q-value estimation is challenging (sparse reward scenario, where the performance of CQL is $3.2 \pm 1.7$), an elevated $\eta$ exacerbates the training process, leading to diminished performance.


\end{document}